\documentclass{article}

\PassOptionsToPackage{numbers, compress}{natbib}
\usepackage[preprint]{neurips_data_2023}
\usepackage[utf8]{inputenc} % allow utf-8 input
\usepackage[T1]{fontenc}    % use 8-bit T1 fonts
\usepackage{hyperref}       % hyperlinks
\usepackage{url}            % simple URL typesetting
\usepackage{booktabs}       % professional-quality tables
\usepackage{amsfonts}       % blackboard math symbols
\usepackage{nicefrac}       % compact symbols for 1/2, etc.
\usepackage{microtype}      % microtypography

\usepackage{graphicx}
\usepackage{amsmath}
\usepackage{amssymb}
\usepackage{booktabs}
\usepackage{colortbl}
\usepackage{multirow}
\usepackage[table,xcdraw]{xcolor}
\usepackage{CJK}
\usepackage{algorithm}
\usepackage{algorithmicx}
\usepackage{algpseudocode}
\usepackage{amsmath}
\usepackage{amssymb}
\usepackage[super]{nth}
\floatname{algorithm}{Algorithm}

\usepackage[skip=1ex]{caption}
\usepackage{tabularx}
\newcolumntype{C}{>{\centering\arraybackslash}X}
\hypersetup{hidelinks} 
\usepackage{xspace}
\usepackage{makecell}
\usepackage{wrapfig}
\usepackage{fontawesome}
\usepackage{xcolor}

\newcommand{\eg}{\textit{e.g.}}

\usepackage{footnote}

\title{RVD: A Handheld Device-Based Fundus Video Dataset for Retinal Vessel Segmentation}

\author{
MD Wahiduzzaman Khan$^{1,2}\thanks{Equal contribution}$ \And Hongwei Sheng$^{1,2*}$ \And Hu Zhang$^{1*}$ \And Heming Du$^{1,3}$ \And Sen Wang$^{1}$ \And
Minas Theodore Coroneo$^{4}$ \And Farshid Hajati$^{5}$ \And Sahar Shariflou$^{2}$ \And Michael Kalloniatis$^{6}$ \And Jack Phu$^{4}$ \And Ashish Agar$^{4}$\And Zi Huang$^{1}$\And Mojtaba Golzan$^{2}\thanks{Corresponding author}$\And Xin Yu$^{1\dagger}$ \\
$^{1}$The University of Queensland, Australia \\
$^{2}$University of Technology Sydney, Australia \\
$^{3}$Australian National University, Australia \\
$^{4}$University of New South Wales, Australia \\
$^{5}$Victoria University, Australia \\
$^{6}$Deakin University, Australia \\
\texttt{arnobk511@gmail.com},~\texttt{mojtaba.golzan@uts.edu.au},~\texttt{xin.yu@uq.edu.au}
}

\begin{document}

\maketitle

\begin{abstract}
Retinal vessel segmentation is generally grounded in image-based datasets collected with bench-top devices. 
The static images naturally lose the dynamic characteristics of retina fluctuation, resulting in diminished dataset richness, and the usage of bench-top devices further restricts dataset scalability due to its limited accessibility. Considering these limitations, we introduce the first video-based retinal dataset by employing handheld devices for data acquisition. The dataset comprises 635 smartphone-based fundus videos collected from four different clinics, involving 415 patients from 50 to 75 years old. It delivers comprehensive and precise annotations of retinal structures in both spatial and temporal dimensions, aiming to advance the landscape of vasculature segmentation. Specifically, the dataset provides three levels of spatial annotations: binary vessel masks for overall retinal structure delineation, general vein-artery masks for distinguishing the vein and artery, and fine-grained vein-artery masks for further characterizing the granularities of each artery and vein. In addition, the dataset offers temporal annotations that capture the vessel pulsation characteristics, assisting in detecting ocular diseases that require fine-grained recognition of hemodynamic fluctuation. In application, our dataset exhibits a significant domain shift with respect to data captured by bench-top devices, thus posing great challenges to existing methods. Thanks to rich annotations and data scales, our dataset potentially paves the path for more advanced retinal analysis and accurate disease diagnosis. In the experiments, we provide evaluation metrics and benchmark results on our dataset, reflecting both the potential and challenges it offers for vessel segmentation tasks. We hope this challenging dataset would significantly contribute to the development of eye disease diagnosis and early prevention. The dataset is available at~\href{https://uq-cvlab.github.io/Retinal-Video-Dataset/}{\faGithub~{RVD}}.
\end{abstract}

\section{Introduction}
The observation of the retinal vasculature patterns serves as a reliable approach to tracking the morphological changes of eyes over time. 
These morphological changes have been found to be closely associated with a spectrum of ocular diseases, \eg, diabetic retinopathy, age-related macular degeneration, and glaucoma~\cite{yu2014functional,cheung2017retinal}. 
Retinal vessel segmentation aims to provide pixel-level extraction of the visible vasculature from a fundus image~\cite{niemeijer2004comparative}.
It is the initial yet fundamental step in objectively assessing vasculature in fundus images and quantitatively interpreting associated morphometrics. Thus, this task plays a pivotal role in understanding and diagnosing ocular diseases.

\begin{figure}[t]
\centering
\includegraphics[width=0.9\textwidth]{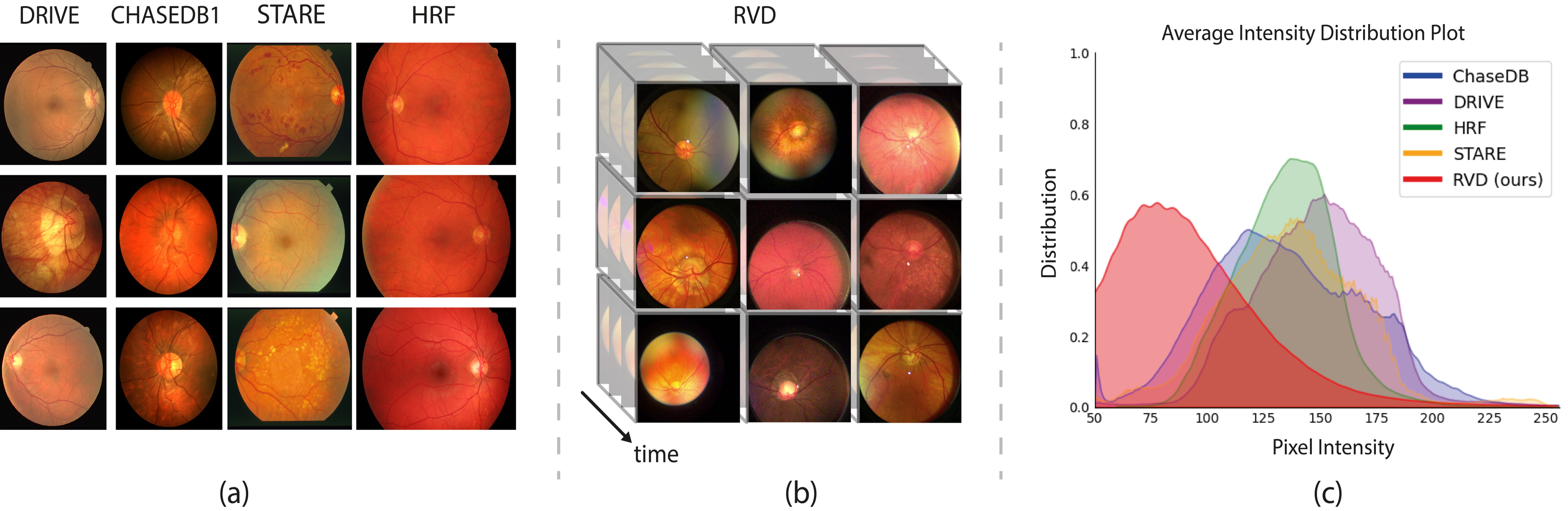}
\caption{ \textbf{(a)} Samples from existing image based retinal vessel datasets: DRIVE~\cite{staal2004ridge}, STARE~\cite{soares2006retinal}, HRF~\cite{kohler2013automatic}, and CHASE\_DB1~\cite{fraz2012ensemble}. 
\textbf{(b)} Video samples from our retinal vessel dataset. 
%highlighting the distinct characteristics of our data.
Different from existing image-based datasets, our dataset captures continuous changes in retinal vessels and facilitates the analysis of vessel dynamics in the retina.
~\textbf{(c)} The intensity distributions of our dataset and existing ones. The differences imply the domain gaps between our dataset and existing ones. 
% \XY{This image is not right}
}
% \vspace{-2em}
\label{fig: some examples}
\end{figure}

Existing methods for retinal vessel segmentation are designed on image-based datasets~\cite{kohler2013automatic, staal2004ridge, soares2006retinal, fraz2012ensemble}, as shown in Fig.~\ref{fig: some examples} (a).
Although these datasets have contributed valuable vessel annotations for studying retinal segmentation, the static nature of images inherently limits their ability to portray dynamic retinal characteristics, \eg, vessel pulsations. % and multi-view
These dynamic phenomena play a vital role in facilitating comprehensive and in-depth understanding of retinal functionality and vasculature structure. %~\cite{}. 
Moreover, image-based datasets are captured by expensive bench-top ophthalmic equipment, which is operated by professionally trained clinicians~\cite{RITE,jin2022fives}. 
Such requirements potentially limit the scale of the datasets and data diversity, thereby adversely affecting the generalization ability of the models trained on these datasets.

% (RVD), a collection of 635 smartphone-based videos with detailed vessel annotation. These videos are45
% recorded from four clinics, including patients from 50 to 75 years old. 
In recent years, advances in imaging technology have enabled the usage of smartphone-based devices for retinal observation~\cite{wintergerst2020diabetic,iqbal2021smartphone}. They offer better flexibility and portability, allowing for scalable data collection.
In this paper, we introduce the first video-based retinal vessel dataset (RVD), a collection of 635 smartphone-based videos with detailed vessel annotation. These videos are recorded from four clinics, including patients from 50 to 75 years old.
% The adopted video modality and the use of portable devices for data acquisition collectively overcome the limitations of existing datasets.
Some examples of our dataset are shown in Fig.~\ref{fig: some examples} (b). 
The sequential frames capture the continuous changes in retinal vessels and thus significantly facilitate the analysis of subtle fluctuations in the retinal structure.
Therefore, the use of portable devices for data acquisition and the provision of the video modality remarkably overcome the limitations of existing datasets.

The annotations provided in our dataset span two dimensions: spatial and temporal. %
In the spatial dimension, we offer three distinct levels of annotations: binary vessel masks, general vein-artery masks, and fine-grained vein-artery masks. 
Each kind of annotation is tailored to specific clinical purposes. 
Specifically, for the binary vessel masks, we identify the sharpest and most representative frame from the video clip and generate binary masks representing the skeletal structure of the vessels. 
This mask primarily targets the holistic vessel structure but neglects the difference between arteries and veins. 
For general vein-artery masks, we differentiate veins and arteries based on their respective vessel calibres and generate separate masks for them respectively. 
Lastly, in contrast to the general differentiation between arteries and veins, the fine-grained vein-artery masks further divide each retinal artery and vein into sections based on a set of pre-defined vessel widths. 
We thus generate eight different vein-artery masks for each sample and these masks precisely reflect the granularities of retinal vessels. These sophisticated masks are highly demanded when detecting ocular diseases~\cite{9398752,CHALAKKAL202059}.%, \eg, xx. 
 % such as glaucoma, diabetic retinopathy, and macular edema
 
%by providing sequential frames in video clips and the usage of portable devices. that  
%capture the continuous changes of retinal vessels and facilitate the analysis of subtle fluctuations in the retinal structure.
%These videos involve patients from four clinics, with an age from 50 to 75 years old.

In the temporal dimension, we enrich our dataset with annotations of the complex dynamics of retinal vasculature. 
For each video, we focus on the optic disk regions where the retinal vessel fluctuation normally occurs. We then select and annotate frames with the maximal and minimal pulse widths as well as label the existence of spontaneous retinal venous pulsations (SVP). 
The existence and extent of vessel changes signify vascular pulsations and cranial pressure-related alterations.
Clinically, the signals of pulsation facilitate the detection of abnormalities in retinal vessels, while precise identification of pressure-related alterations aids in detecting %and monitoring 
temporally-dependent ocular diseases.
%, which can hardly be provided in existing datasets. 
Our integration of temporal annotations thus %enhances the richness of the dataset and
increases its potential for ocular disease diagnosis. 

The distinction between smartphone-based and benchtop devices and data modality differences result in domain gaps, as illustrated in Fig.~\ref{fig: some examples}. 
Furthermore, since our data are collected by handheld devices in clinics, our dataset also involves more realistic factors, \eg, the operations of the clinicians, surrounding illumination conditions, and eye movements of patients during video capture. Consequently, our dataset presents more challenges for existing vessel segmentation methods. More importantly, the large number of training samples and detailed annotations in our dataset will likely pave the way toward more advanced yet portable retinal analysis and more accurate disease diagnosis. 
In the experiments, we delve into an in-depth analysis of our dataset and provide benchmark results of different tasks on our newly curated dataset.
The main contributions of our paper are summarized as follows:
% \vspace{-0.8em}
\begin{itemize}
    \item \textbf{Dataset construction}: We construct a new video-based retinal vessel dataset (RVD) with rich spatial and temporal annotations for vessel segmentation tasks. To the best of our knowledge, RVD is the first mobile-device based dataset for retinal vessel segmentation.
    \item \textbf{Three-level spatial annotations}: Our dataset introduces three levels of annotations in spatial, comprising binary vessel masks, general vein-artery masks, and fine-grained vein-artery masks. The hierarchical and diverse descriptions of spatial annotations enable us to better analyze the vessel structure.
    %These annotations together considerably extend the application of our dataset beyond the existing datasets. 
    %constructed by considering the varying complexity of vessel structure considerably extends the application of our dataset beyond the existing datasets.
    %a unique contribution to the width-based categorization of retinal vessels, accompanied by the vein-artery annotation system. By categorizing vessels based on widths, we enable a more detailed understanding of the structural characteristics of different vessel types. 
    \item \textbf{Temporal annotations}: Our dataset also provides temporal annotations of spontaneous retinal venous pulsations (SVP) to reveal the dynamic changes in retinal vessels. This enables the assessment of pulsatile variations in retinal vessels. %and intraocular pressure.

    %{Need changes below}
    \item \textbf{Benchmarking}: We investigate the gap between our dataset and previous retinal datasets by assessing the performance of several state-of-the-art methods. The experimental results will shed some light on mobile-device based retinal vessel segmentation.
    % the challenges and opportunities associated with our proposed dataset in the area of
\end{itemize} 

\section{Related Work}
% \vspace{-0.5em}
\textbf{Existing Retinal Datasets:}
In the realm of retinal vessel segmentation, various retinal vessel datasets have been proposed. Existing datasets can be roughly categorized into two streams: binary vessel based ones and artery-vein based ones. Among the datasets with binary vessel masks, DRIVE~\cite{staal2004ridge}, STARE~\cite{soares2006retinal}, HRF~\cite{kohler2013automatic}, and CHASE\_DB1~\cite{fraz2012ensemble} have emerged as the most frequently used datasets. In fact, each of these datasets only comprises dozens of images. For example, DRIVE consists of 40 images captured in the 45-degree field of view, with an image size of 584$\times$565 pixels. %STARE comprises 20 retinal images (700$\times$605 pixels each) and all the images are captured at a 35-degree field of view. %and provided by the University of California.
Besides, DRiDB~\cite{prentavsic2013diabetic}, ARIA~\cite{farnell2008enhancement}, IOSTAR~\cite{zhang2016robust}, and RC-SLO~\cite{abbasi2015biologically} are another publicly available datasets for retinal vessel segmentation. However, they are less used in recent years considering their data quality and maintenance. Recently, the FIVES dataset has been introduced~\cite{jin2022fives}, with data distributed across four categories: normal retinas, retinas affected by Diabetic Retinopathy, Glaucoma, and Age-related Macular Degeneration. It comprises 800 retinal images.
% with a resolution of 2,048$\times$2,048 pixels. 

Regarding the datasets with artery-vein masks, %RITE~\cite{RITE} and 
RITE~\cite{RITE}, AV-DRIVE~\cite{qureshi2013manually}, INSPIRE-AVR~\cite{INSPIRE_AVR}, and WIDE~\cite{estrada2015retinal} are the available ones. The AV-DRIVE dataset, derived from DRIVE, consists of 40 images and offers separate ground truth masks for arteries and veins. The INSPIRE-AVR is an independently constructed dataset with artery-vein ground truth masks. It consists of 40 color images in total. The WIDE dataset provides 30 scanning laser ophthalmoscope (SLO) images.
% with a resolution of 3,900$\times$3,072 pixels.
%along with their artery-vein ground truth labels annotated in~\cite{estrada2015retinal}.

In contrast to existing datasets which are collected with cumbersome bench-top devices and are composed of static images, our dataset is constructed with portable handheld devices and is video-based. Our dataset preserves the dynamic characteristics of vessels. Besides, existing datasets typically provide only one type of annotation for a specific research purpose, whereas our dataset offers annotations in both spatial and temporal dimensions. The spatial annotations include binary vessel masks, general vein-artery masks, and fine-grained vein-artery masks, respectively. The temporal annotations reveal the state of SVP, an important signal for diagnosing various diseases. 

\textbf{Methods for Retinal Vessel Segmentation:}
In the past years, a variety of methods have been developed for retinal vessel segmentation. Traditional methods mainly depend on handcrafted features~\cite{barkana2017performance, dash2017thresholding, zhou2020new, tchinda2021retinal, ramos2021efficient, hashemzadeh2019retinal,dash2020enhancing,khan2022width}, which are less discriminative and effective~\cite{mookiah2021review}. With the unprecedented breakthroughs of deep neural networks (DNNs) in the image classification, detection, and segmentation tasks, researchers have explored the potential of DNNs in retinal vessel segmentation~\cite{fan2016automated, liskowski2016segmenting,khalaf2016convolutional,wu2016deep, zhou2017improving}. %In \cite{samuel2021vssc}, VSSC Net integrates vessel-specific convolutional blocks and skip chain convolutional layers to effectively fuse intermediate features from VGG-16~\cite{simonyan2014very}, thereby enhancing the extraction of retinal vessels.
%In~\cite{soomro2017boosting}, they focus on the alleviation of low contrast, uneven illumination, and noise visual complexities in given images by morphology operations and principal component analysis. 
Many works~\cite{atli2021sine, li2020fundus,soomro2019strided,oliveira2018retinal,jiang2018retinal,dasgupta2017fully} adopt fully convolutional networks~\cite{long2015fully}  to produce more accurate segmentation of retinal images by combining semantic information from deep layers with appearance information from shallow layers.
% Considering the resilience of U-Net~\cite{ronneberger2015u} to smaller datasets, %and their ability to generate outputs of the same resolution as the input, 
Several works have focused on modifying the U-Net structure~\cite{sule2020enhanced,sathananthavathi2021encoder,zhang2020befd} for vessel segmentation. \cite{zhang2018deep} first introduces the residual connection into U-net to detect vessels. This idea has been adopted in later studies~\cite{alom2018recurrent,li2019residual,gegundez2021new, wei2021genetic}.
%In~\cite{wang2019dual}, a novel Dual Encoding U-Net (DEU-Net) has been proposed. It consists of two encoders: a spatial one with a large kernel to preserve the spatial information and a context one with a multiscale convolution block to capture more semantic information. %To construct Genetic U-Net~\cite{wei2021genetic}, a condensed but flexible search space is first devised based on a U-shaped encoder-decoder. Then, an improved genetic algorithm is used to identify architectures that exhibit superior performance in the search space.
\cite{xu2022local} introduces the local-region and cross-dataset contrastive learning losses in training to explore a more powerful feature embedding space. Besides, several other methods employ various networks and strategies for retinal vessel segmentation, such as generative and adversarial networks~\cite{lahiri2020retinal,son2019towards},  %auto-encoders~\cite{adarsh2020dense}, 
ensemble learning~\cite{maji2016ensemble,tang2019multi,liu2019unsupervised}, and graph convolutional network~\cite{shin2019deep}.

%Some other works either use the attention gate~\cite{li2019residual} or weighted attention gate~\cite{lian2019global} to decrease the interference of background features.
%To increase the depth of the network and prevent gradient vanishing, the residual connections between layers are considered in~\cite{zhang2018deep,alom2018recurrent,li2019residual,gegundez2021new}.
%Works~\cite{cheng2020retinal,wang2019dense} instead use dense U-Net to avoid the generation of redundant activation maps and prevent the loss of vessel details, achieving better predictions with minimal parameters and computational cost. \cite{jin2019dunet} introduces deformable receptive fields in U-net to address the diversity in the size, shape, and scale of retinal vessels. %Based on the U-net structure, extensive works have been undertaken to investigate the potential of generative and adversarial networks (GAN) for retinal vessel segmentation~\cite{lahiri2020retinal}. For example, Retina-GAN is proposed  in~\cite{son2019towards} and it uses the U-Net as the generator and Image GAN as the discriminator. Compared to the original U-Net, Retina-GAN tends to exhibit superior preservation of fine retinal blood vessels on the DRIVE and STARE datasets compared to the original U-Net. 

The aforementioned methods are mainly conducted on datasets DRIVE, STARE, CHASE\_DB1, and HRF, with binary vessel masks as supervision. Thanks to INSPIRE-AVR, AV-DRIVE, and WIDE, many works have been proposed to distinguish artery and vein~\cite{dashtbozorg2013automatic,estrada2015retinal,zhao2019retinal,mendoncca2013automatic}. %\cite{srinidhi2019automated} exploits both local and global information to distinguish arteries and veins. The local information is used to disentangle the complex vascular tree into multiple subtrees, and global information is used to label these vessel subtrees into arteries and veins. 
In~\cite{ma2019multi}, a multi-task deep neural network with spatial activation is proposed. The constructed network is able to segment full retinal vessels, arteries, and veins simultaneously.
More recently, transformer based models, \eg, ViT~\cite{dosovitskiy2020image}, Swin transformer~\cite{liu2021Swin}, and Mask2Former~\cite{cheng2022masked}, have been proposed. These models have demonstrated their superior performance in capturing visual concepts and become popular backbones in visual understanding tasks~\cite{naseer2021intriguing}. 
%They are now widely used.
We thus choose these models in the experiments to study the characteristics of our proposed dataset.

% \vspace{-1.0em}
\section{Our Proposed RVD} % what why how
% \vspace{-1.0em}
In this section, we first describe our data collection process and data sources. 
Concerning privacy and ethic, we perform this study in accordance with the guidelines of the Tenets of Helsinki. 
Written consent was obtained from all participants prior to any data collection, and all examination protocols adhered to the tenets of the Declaration of Helsinki. 
Once clinical data have been collected, we need to clean and pre-process the data in order to facilitate clinicians' annotations and neural network training. 
In our work, the annotations are provided by professionally well-trained clinicians, and they have been asked to not only annotate conventional spatial segmentation masks but also temporal segmentation masks for dynamic biomarkers, such as Spontaneous retinal Venous Pulsations (SVPs).
Last, we will introduce the data split and relevant tasks that are supported by our dataset. 

% This is followed by an explanation of the necessary data cleaning and preprocessing steps. Subsequently, we elucidate the annotations and potential applications of our dataset. Finally, we detail the protocols implemented for the division of the dataset. Concerning privacy and ethic, we perform this study in accordance with the guidelines of the Tenets of Helsinki and approved by the University of Technology Sydney's Human Research Ethics Committee. Written consent was obtained from all participants prior to any data collection, and all examination protocols adhered to the tenants of the Declaration of Helsinki.

\begin{table}[t]
\centering
\scriptsize % 
\caption{Comparisons of different retinal vessel segmentation datasets. ``Num'' denotes the number of annotated image frames. 
% number of annotations.
}
\setlength{\tabcolsep}{1.5 mm}{
\begin{tabular}[width=\linewidth]{c|c|c|c|c|c|c}
% \footnotesize % 
\toprule
\textbf{Dataset} & \textbf{Resolution} & \textbf{Modality} & \textbf{Device} & \textbf{Num} & \textbf{Dimension} & \textbf{Annotation type} \\
\midrule
STARE~\cite{soares2006retinal} & 605$\times$700 & Image & Benchtop  & 20 & Spatial & Binary \\
DRIVE~\cite{staal2004ridge} & 768$\times$584 & Image & Benchtop  & 40 & Spatial & Binary \\
ARIA~\cite{farnell2008enhancement} & 576$\times$768 & Image & Benchtop  & 161 & Spatial & Binary \\
CHASEDB1~\cite{fraz2012ensemble} & 990$\times$960 & Image & Benchtop & 28 & Spatial & Binary \\
INSPIRE-AVR~\cite{INSPIRE_AVR}  & 2392$\times$2048 & Image & Benchtop  & 40 & Spatial & Multi-class \\
HRF~\cite{kohler2013automatic}  & 3304$\times$2336 & Image & Benchtop  & 45 & Spatial & Binary \\
RITE~\cite{RITE} & 768$\times$584 & Image & Benchtop  & 40 & Spatial & Multi-class \\
FIVES~\cite{jin2022fives}  & 2048$\times$2048 & Image & Benchtop & 800 & Spatial & Binary \\
RAVIR~\cite{RAVIR}  & 768$\times$768 & Image & IR  Laser & 42 & Spatial & Multi-class \\
\rowcolor[HTML]{EFEFEF}
\textbf{RVD (ours)} & \textbf{1800$\times$1800} & \textbf{Video} & \textbf{Hand-held} & \textbf{1,270} & \textbf{\makecell[c]{Spatial \\+ Temporal}} & \textbf{Multi-class} \\
\bottomrule
\end{tabular}}
% \vspace{-2.0em}
\label{tab:dataset_comparison}
\end{table}
% all benchtop before, di they mentioned?
% \vspace{-1.0em}
\subsection{Data Collection}
% \vspace{-0.75em}
% data sources: diversity disease ... data handheld device, smartphone...bias healthy/not
For data collection, the employed hand-held fundus imaging devices are constructed by connecting a smartphone to the fundus camera lens. Then, clinicians are trained to operate the hand-held devices to examine patients' retinas while collecting fundus videos.
These participants are fully aware of data collection when they undergo their annual medical examinations.
With the help of clinicians, a total of 415 patients from four different clinics participate in the data collection process.
As data are collected in different clinics over the past five years, the employed smartphones are different, thus increasing the diversity of data sources. 
More specifically, 264 males and 151 females are included here. 
Their ages range from 50 to 75. 
People of these ages are commonly considered to be at high risk for eye-related diseases, such as glaucoma and hypertension~\cite{klein2011prevalence}.
Our dataset involves both videos recorded from healthy eyes and videos from eyes with ocular diseases.
% The device allows us to steadily capture the fundus videos with minimal environmental interference.
% We then deliver the devices to clinics and ask professional clinicians there to help collect fundus videos.% using their own smartphones.% from annual checking patients.
% The use of different smartphones introduces domain gaps in the dataset.
% All the videos in the dataset are collected with the designed hand-held devices.
% 

% During the collection process, 

% As we do not specifically record data for a particular disease, the eye conditions we recorded exhibit both healthy and non-healthy status, ensuring diversity in our dataset.
%We randomly choose the participants when they do. Our dataset We randomly select the participants and xx . Little bias, for example, the preference for 

During the collection, one eye of each patient is recorded at a time.
In this manner, at least one fundus video of each patient has been recorded and some participate multiple times in video recording.
As a result, a total of 635 RGB videos have been captured.
% Each video represents the condition of one eye of the participants.
All captured videos have a frame rate of 25 frames per second, with the duration varying between 2 to 30 seconds.
The total number of frames in our dataset is over 130,000.
This collection process ensures the generality and diversity of our dataset for retinal vessel analysis. 
The detailed information of our dataset is shown in Table~\ref{tab:dataset_comparison} and some examples could be found in Fig.~\ref{fig: some examples} (b).  

%comprehensive analysis and evaluation in 
%To advance research in fundus imaging, we have collected a dataset of smartphone fundus videos from multiple clinics. The dataset encompasses 635 RGB videos collected from four different clinics,
% across Sydney, Australia, involving 415 patients. More specifically, the dataset includes 264 male participants with age of 65±11 years. This age group is considered a high-risk population for both glaucoma and hypertension.
%The video fraes as shown in Fig. \ref{dataset}, are saved in a rectangle shape with a side length of 1800 pixels.
% \vspace{-2em}
%\wei{figure lack}
% \vspace{-1.0em}
\subsection{Data Cleaning and Preprocessing}
% \vspace{-0.75em}
Although we have tried our best to minimize environmental interference during collection, the original videos still exhibit various noise, %as shown in Fig.~\ref{xxx}, 
such as video jittering and motion blur.
%, imaging noise, and light interference.
% which prove more severe when compared to existing datasets acquired using benchtop devices, as depicted in Fig.~\ref{}(a). 
Such noise will severely degrade the quality of collected videos and impose more difficulties in annotations. Hence, we eliminate the noise in the footage to improve the quality of our dataset and facilitate annotations.

%As depicted in Fig.\ref{}, the primitive videos collected could contain diverse noise, such as the missing optic disc regions (ODR) or blurring imposed by %the inherent complexities 
%of real-world clinical conditions. Such noise will severely degrade the quality of collected videos and impose more difficulties in annotations. We thus propose a set of methods to identify and eliminate the noise in the footage. 

%\textbf{Remove ODR-invisible Images.} % rewrite this with a good reason such as for a better vessel display
% ODR serves as the entry point for the major blood vessels that supply the retina.~\cite{}

%As the blood vessels passing through ODR play a crucial role in retina functioning \cite{}, we first remove the video segments without presenting ODR.
\textbf{Data cleaning: } Considering that blood vessel dynamics mostly appear in ODR, we remove the video segments without ODR and ensure the existence of ODR in all videos.
% since some diseases, such as Spontaneous Venous Pulsation (SVP), demand continuous detection of ODR.
% Given a video, we first utilize a detection network~\cite{} to locate the ODR in each frame. 
To this end, we employ an ODR detection method to localize ODR. Specifically, we label ODR regions by bounding-boxes, and for each video, we only annotate one frame per 25 frames (\emph{i.e.}, 1 second), similar to \cite{hongwei}. Then, we leverage the labeled ODR as supervision to train the Faster-RCNN detection network~\cite{ren2015faster}. After that, the remaining frames are labeled by the trained Faster-RCNN. Manual check is also conducted to modify erroneous detection results by annotators.
We only select video segments in the dataset if their ODRs are detectable in a minimum of 30 continuous frames. Such operations help maintain the overall quality of our video data.
%Then in each video, we employ a Faster-RCNN detection network~\cite{} to scan the optic disc region (ODR) in each frame of the video.
%We then select video segments where ODR is detected in at least 30 continuous frames (\emph{i.e.}, 1-second duration) to ensure the quality of our video data.
%To confirm the existence of SVP, ODR needs to be continuously detected as SVP is a dynamic process of the blood vessels around it.

To further improve the data quality, we leverage the optical flow to pinpoint frames with a high level of blur. 
Optical flow captures the spatial alterations between distinct frames, and thus it could serve as an indicator of spatial sharpness. 
Frames with large optical flow are subsequently discarded, as they likely correspond to instances of blurring. Similar to ODR detection, annotators (non-experts) also manually remove frames that undergo severe blur but have not been spotted by optical flow.

%Since some diseases, such as Spontaneous Venous Pulsation (SVP), are a dynamic process of the blood vessels around the optic disc region (ODR), their detection actually demands continuous detection of ODR. Considering this case, we remove the video without good ODR. We first utilize a detection network to locate the ODR in each frame. We then select video segments where ODR is detected in at least 30 continuous frames (\emph{i.e.}, 1-second duration) to ensure the quality of our data.
%To confirm the existence of SVP, ODR needs to be continuously detected as SVP is a dynamic process of the blood vessels around it.
%\textbf{Remove Blur Images.} 
% We then filter the blur frames based on the variation of detected ODR bounding boxes. 
%The blurring is normally caused by excessive shaking. 
%address the jitter by 
%calculating the deviations of the detected ODR bounding box between different frames. In addition,  
% Our method then utilizes bounding boxes to eliminate the jitters in the footage. %However, since the fluctuation around the ODR is small compared to the sizes of bounding boxes, our method still obtains the approximate location of the ODR in each frame, enabling further screening extreme jitters.
\textbf{Data Preprocessing: } % SVP-ODR-based-stablization, see Q in 3.3.2.2 
In retinal vessel segmentation, ODR and its surrounding area are the most representative regions of the eye and provide extensive details about retinal vessels. However, the inherent ocular movement results in varying ODR positions across different frames.
% , which cannot be addressed by the cleaning methods above. 
Such variations can impede the precise observation and annotation of SVP by clinicians. To tackle this issue, we employ the template matching algorithm~\cite{brunelli2009template} to stabilize the ODRs across video segments, ensuring a consistent ODR placement and a fixed field of view across frames. This facilitates human observation and machine perception of dynamic changes surrounding the ODR, thus greatly enhancing annotations and clinical diagnosis.

\begin{figure}[t]
\centering
% \scalebox{1.12}
\includegraphics[width=\linewidth]{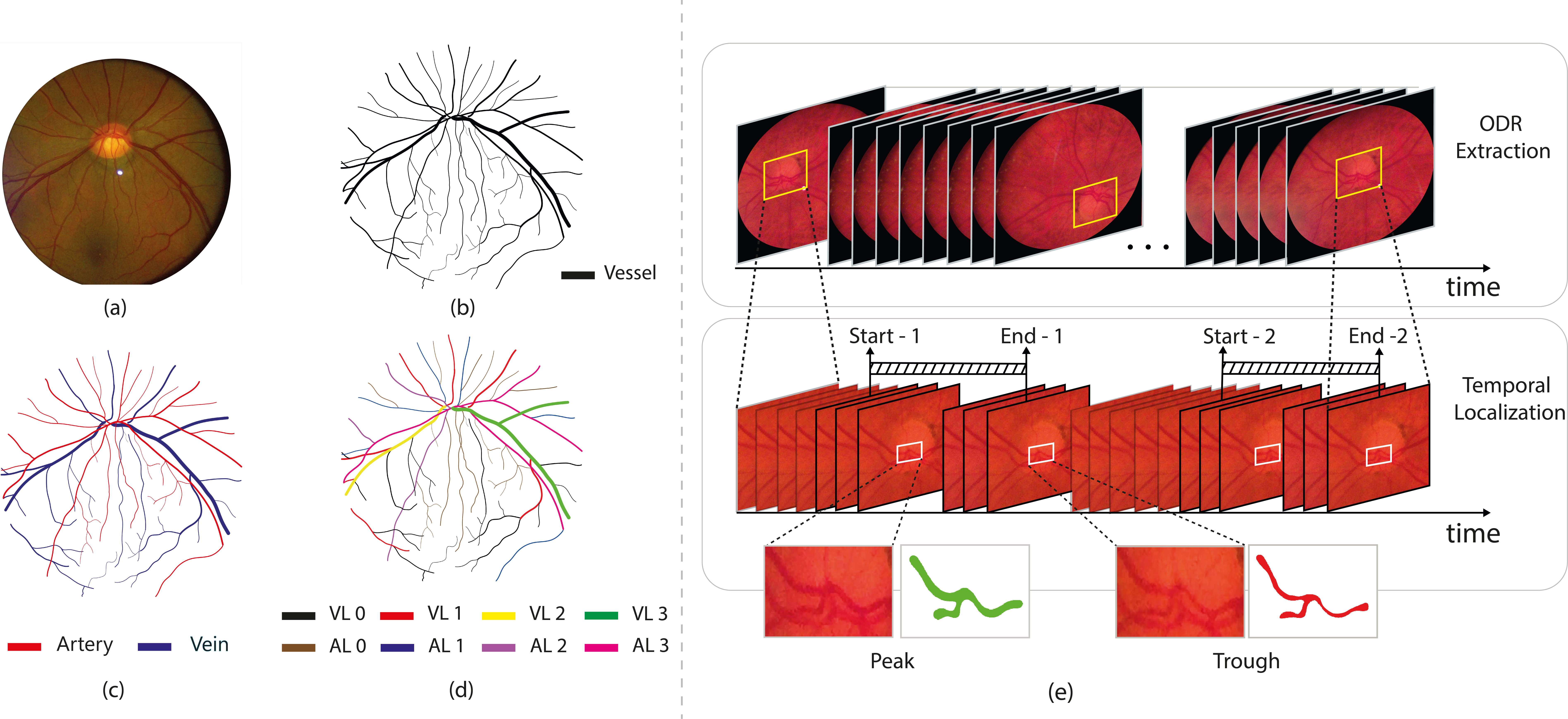}
% \vspace{-1.0em}
\caption{
% Illustration of the multi-grained annotations for our retinal video dataset.
% % retinal image showcasing different layers of annotations. 
% \textbf{(a)} Original retinal video frame, \textbf{(b)} Annotated binary mask, 
% \textbf{(c)} Annotation of veins and arteries, 
% \textbf{(d)} Fine-grained artery-vein annotation based on vessel widths (VL: vein width level, AL: artery width level, where level 0 represents the thinnest one while level 3 indicates the thickest one). 
% \textbf{(e)} Temporal annotations: 
\textbf{Left:} Illustration of our multi-grained segmentation annotations. For each given fundus image \textbf{(a)}, we provide three different kinds of segmentation masks including a conventional binary mask \textbf{(b)}, a general artery-vein mask \textbf{(c)} and a fine-grained artery-vein mask \textbf{(d)} (VL: vein width level, AL: artery width level, the numbers (0 to 3) indicate four increasing width levels). 
\textbf{Right}: Overview of the temporal annotations~\textbf{(e)}, including ODR locations, presence and absence of SVP, temporal localization of SVP, and ``peak'' and ``trough'' of SVP.
% Our pre-processed fundus videos can make the most dynamic feature~(\ie, {SVP}) more evident. We then annotate the status of SVP appearance and the endpoints of SVP appearing intervals in each video. The 'peak' and 'trough' indicating maximum and minimum diameters of pulsating vessels are specifically annotated as segmentation maps for potential medical research.
}
% indicating time-related changes within the retinal videos.}
% \vspace{-1.5em}
\label{fig: annotation examples}
\end{figure}

%do we really need to mention the 3200*1800 videos?

%Finally, we crop and remove redundant parts outside of the ODR to further emphasize the SVPs.

% \vspace{-0.75em}
\subsection{Data Annotation}
% \vspace{-0.75em}
After data cleaning and preprocessing of the initial videos, we detail the annotation process and provide the annotations for both spatial vessel segmentation and temporal SVP localization and classification tasks. To ensure the annotation quality, six clinicians are involved in this process. %\hu{The annotation results are also illustrated in Fig.~\ref{fig: annotation examples}}
% \vspace{-0.75em}
\subsubsection{Spatial segmentation}
% \vspace{-0.75em}
%The overall process of our annotation can be divided into two sections: (1) keyframe selection; (2) skeleton and borderline detection.
%\paragraph{Spatial Key Frame Selection:}
% and generate the annotation for the rest frames automatically.
To mitigate the redundancy of annotating similar video frames, we select two keyframes from each video and generate subsequent annotations for them.
Here, we adopt a three-fold strategy to identify the most representative frame from a video. %as the first keyframe.
\textbf{(1)} A frame contains the most vessels in a video; 
\textbf{(2)} A frame can cover ODR, fovea, and macula regions since these regions pathologically have a significant number of capillaries;
\textbf{(3)} After one frame is selected, we choose another frame that not only has visible ODR with high-density vessels but also exhibits maximum spatial distance between its ODR and the previously selected ones. In this way, the selected frames will cover most of the retinal vasculature and any pathological regions of the retina.
We then elaborate on creating three types of spatial annotations: binary vessel masks, general vein-artery masks, and fine-grained vein-artery masks, as follows:
\textbf{Binary vessel masks:} 
% \WKA{May need to discuss for overlaps}
To generate the binary vessel masks, we adopt a similar method proposed in~\cite{9284503}.
For each frame, we first draft a centerline-level annotation using the ImageJ software~\cite{schneider2012nih} and generate the delineation of vessel boundaries to obtain the main structure of vessels.
Then we employ our experts to manually refine the structure by correcting the boundaries and improving the details of small capillaries. We can obtain the binary vessel masks by assigning the label to the refined structure (see Fig.~\ref{fig: annotation examples} (b)).  
%By filling in the empty gaps between the boundaries and centerline, we can obtain a complete mask annotation for binary vessel segmentation, as illustrated in Fig.~\ref{fig: annotation examples} (b).
% An expert clinician was initially engaged to grade microvascular segments at the pixel level. 
% Due to the difficulty in accurately assessing the diameters of small capillaries near the macula region, the clinician is instructed to grade the small capillaries at the centerline level. 
% When the borderlines of vessels are constructed, we fill the area within the borderlines with predefined mask values to obtain the binary mask.

\textbf{General vein-artery masks:}
Many intracranial vascular diseases are found to be related to retinal vessels and affect the arteries and veins differently~\cite{abramoff2010retinal}.
Thus, distinguishing between the retinal artery and vein plays a critical role in the clinical biomarker study of how various systemic and cardiovascular diseases affect the retinal vessels. %In clinical settings, different retinal diseases have varying impacts and manifestations on arteries and veins.
In practice, the arteries and veins can be distinguished based on their difference in three aspects: color, light reflection, and calibres.
The veins generally have a darker color than arteries and show a smaller central light reflex.
Meanwhile, the veins are also wider than adjacent arteries. Then, %Once we attain the binary vessel masks, 
clinicians only need to assign labels (\emph{i.e.}, vein and artery) to the vessels and obtain the vein-artery masks, as shown in Fig.~\ref{fig: annotation examples} (c).
%Once we attain the binary vessel masks, clinicians only need to assign labels (\emph{i.e.}, vein and artery) to the masks.
% All the annotations have been manually checked by the clinicians.
% Based on the separated arteries and veins, we combine the vessel borderlines obtained in binary vessel masks and assign different masks for arteries and veins.

%Original artery-vein recognition is a significant challenge task, no matter whether it is performed manually~\cite{hu2021automatic} or automatically~\cite{hu2021automatic}.
% The walls of a vein are thinner than an artery

%The phenomenon of arteriovenous nicking can only be seen when an arterial vessel is crossing over a venous one.

%We use the criterion of the retinal pictures' branching pattern and carried out three procedures. Ophthalmologists are initially tasked with locating the retina and veins within the optic disc. Afterward, we trace the retinal branching pattern and conveniently modify the annotation by recording and adjusting the segment points. 
\textbf{Fine-grained vein-artery masks:} 
Vascular morphology holds substantial clinical significance, as alterations in vessel diameters frequently signify the presence of various diseases. For example, {damage of the small retinal vessels could result in diabetic retinopathy}~\cite{kumar2012diabetic}. 
Similarly, glaucoma pathogenesis is postulated to be linked to alterations in the retinal vasculature, such as retinal arteriolar narrowing and decreased fractal dimension~\cite{bmjophth-2016-000032corr1}. 
% , a hypothesis backed by numerous clinical and population-based studies
%\XY{wrong english}
% To elucidate these associations further, we have developed a methodology that categorizes individual vessel segments relative to the central arteries and veins.
% For example, central retinal artery occlusion typically manifests as exceptionally narrowed retinal arteries.
Despite the clinical importance of such information, existing datasets scarcely provide this type of labels.
Therefore, we consider the morphological characteristics of each artery and vein in our dataset and thus provide fine-grained vein-artery masks based on the vessel widths.

Specifically, we first measure the vessel diameters automatically via ``Vessel Diameters'' plugin in ImageJ.\footnote{\url{https://imagej.net/software/imagej/}}
Then, we divide the arteries into multiple small vessel segments based on the diameters of the vessels. 
% Initially, for each image, we depend on the previously generated centerline-level annotation to divide the arteries into multiple small vessel segments. 
% Then, we adopt the  to calculate the diameter of each segment.
Based on the largest diameter among these artery segments, we define four levels of widths according to specific ratios.
More specifically, a vessel segment within the range of 0-25\% of the largest diameter is categorized as level 0. Similarly, levels 1, 2, and 3 correspond to vessel widths in the ranges of 25\%-50\%, 50\%-75\%, and 75\%-100\% of the largest diameter, respectively.
Afterward, we obtain four-class masks for arteries based on vessel widths.
The same operation is also applied to veins. 
After the automatic processing, clinicians will validate the quality of the fine-grained segmentation masks. 
This process ultimately yields eight-class masks for both arteries and veins (see Fig.~\ref{fig: annotation examples} (d)). Those masks significantly enrich the granularity of our dataset.

\subsubsection{Temporal localization}

\textbf{{Existence of SVP:} }
Based on the results of data cleaning and preprocessing, we utilize the stabilized videos and further annotate the dynamic state of vessel pulsations. 
Spontaneous retinal Venous Pulsation (SVP) plays a crucial role as a biomarker in retina assessments.
Specifically, SVP is characterized by rhythmic pulsations evident in the central retinal vein and its branches, typically observable within the optic disc region (ODR) of the retinas.
%SVPs refer to rhythmic pulsations observed in the central retinal vein and its branches, typically occurring within the ODR of the retina.
The absence of SVP holds substantial clinical significance, as it is correlated with certain pathologies. For example, the absence of SVP is associated with progressive glaucoma \cite{understandingretinal}, and it is indicative of increased intracranial pressure~\cite{iih}. Considering the needs of the specialty, we invite multiple clinicians to annotate the presence or absence of SVP in each video of our dataset.
% Once the annotation process completes, we obtain ``SVP-present'' and ``SVP-absent'' videos.
{Once the annotation process completes, we obtain 335 ``SVP-present'' videos and 300 ``SVP-absent'' videos respectively.}
This annotation establishes a fundamental task of SVP detection, facilitating further analysis and investigation on the relationship between SVP and eye diseases.

\textbf{Temporal duration of SVP:}
After annotating the existence of SVP in the stabilized videos, some ``SVP-present'' videos may not contain SVP throughout the whole video. This means in some frames SVP is not visible. Using these videos to train an SVP classification model would suffer ambiguity especially when an entire video cannot be fed into a neural network. 
% an inconsistent occurrence of SVP in some videos labeled as , which means the SVP disappears in some frames. This inconsistency can significantly raise the ambiguity in clinical diagnosis~\cite{}. To reveal such a phenomenon and further distinguish the videos, %mitigate this issue, 
Therefore, we further provide temporal emergence annotations of SVP by indicating the starting and ending frames of retinal vessel fluctuation (see Fig.~\ref{fig: annotation examples} (e)). The detailed duration of SVP serves two purposes: it acts as a valuable signal to improve the performance of SVP detection tasks and concurrently sets a new task for SVP temporal localization.
We obtain videos in three distinct groups: 156 videos containing intermittent SVPs, 179 videos demonstrating persistent SVPs, and the remaining 300 videos without SVPs. 
These temporal annotations allow us to better understand retinal vessel dynamics.

% enhances our understanding of SVP patterns and sets a robust foundation for future applications in diagnosing retinal diseases.
% In training a neural network, the labeling indicating whether retinal vessel fluctuation exists in a given video can significantly raise ambiguity while training.
% By considering the temporal aspect of SVP, we can gain valuable insights into the duration and patterns of these pulsations, leading to improved recognition and localization techniques in related research areas.
% Upon reviewing the stabilized videos, the experts observe that some videos were to be labeled as SVP-positive despite the absence of consistent SVP in these videos. This can significantly raise the ambiguity in SVP existence while training. To mitigate this issue, we further apply temporal labeling of SVP in videos where clinically sampled SVP may not be visually apparent throughout the stabilized clips.

\textbf{``Peak'' and ``Trough'' annotations of SVP:} 
As discussed above, SVP reflects the temporal dilation and contraction in retinal vessels. 
The state with maximal dilation is characterized as ``peak'', whereas those with maximal contraction are termed ``trough''. 
Here, we select frames corresponding to the ``peak'' and ``trough'' states from each ``SVP-present'' video. 
Subsequently, we generate corresponding masks for these selected frames, yielding a total of 670 annotated masks.
This annotation allows us to quantitatively measure the extent of pulsations and occurring positions of vessel pulsations.
% thus enabling a more comprehensive assessment of the patient's current health condition. 

%Additionally, it opens up opportunities to explore the quantitative connections between SVPs and other health indicators.

%Furthermore, we combine the dynamic characteristics of SVP with static spatial segmentation and generate the segmentation of the shifting part on the vessels.
%quantitatively represent the size of the region where the pulsation appears.
%To be specific, with pulsating vessels periodically changing over frames, a complete SVP contains one peak and one trough.
%From each `SVP-present' video, our experts select two frames representing peak and trough respectively, and generate the segmentation maps for the parts where SVPs present.
%Note that the segmentation map of the peak image signifies the maximum size of the shifting part, while the segmentation map of the trough image correspondingly reflects the minimum size.
%We obtain XX annotated maps of shifting parts in total.
%This promises us to quantitatively measure the extent of pulsations, thus enabling a more comprehensive assessment of the patient's current health condition.
%Additionally, it opens up opportunities to explore the quantitative connections between SVPs and other health indicators.

% \vspace{-0.75em}
\subsection{Data Protocols}
% \vspace{-0.5em}
% data split Metrics
\label{subsec:protocol}
% \wei{maybe we write split and metrics again based on tasks, each be like seg task:data split\&metrics}
%The videos in our dataset present slight differences in data distribution since they are collected by four different clinics.  
\textbf{Data split:}
When partitioning our dataset for training and evaluation, we also take into account the similarity of the recorded videos of the same person. 
Specifically, we ensure that videos captured from the same patient are allocated to the same subset during the partitioning process. 
This strategy aims to decrease the similarity between training data and testing data and thus minimizes performance bias. %, which is raised due to the inherent high similarity within data sourced from the same individual.
In practice, we divide the data based on patient IDs.
In this manner, the same patient's videos will not appear simultaneously in both the training and testing sets.
Then, we select 517 videos from the 635 videos for training and validation, and the rest 118 videos are used for testing. 
We also cross-validate a method with three different data splits and will release the dataset and data splits. % train val test 457 60 118
% Each video is accompanied by the six annotations above.
% More detailed information could be found in Appendix. 

%diverse domains and such a phenomenon is closely related to the clinical settings.

%In a nutshell, our dataset will be partitioned considering the significant bilateral symmetry between an individual's eyes and inner domain gaps between different clinics.

%Moreover, considering the substantial similarity and bilateral symmetry observed between the two eyes of an individual, it is imperative to account for this characteristic when spliting our dataset.

%To address this, we ensure that videos collected from the same person are allocated to the same split during the partitioning process. 
%This measure is taken to mitigate the possibility of inflated results caused by the inherent high similarity present within data originating from the same individual.
%In a nutshell, our dataset will be partitioned considering the significant bilateral symmetry between an individual's eyes and inner domain gaps between different clinics. 

\textbf{Metrics:}
Based on our annotations, we can conduct tasks in two major categories.
% To facilitate the evaluation of algorithms on the tasks, we also provide the related metrics here.
% Based on the types of data we annotated, we have established two major categories of benchmarks.
{(1) Retinal vessel segmentation metrics:}
% For annotations based on spatial information, we have established three tasks on retinal vessel segmentation datasets.
% Specifically, we set Binary Segmentation task on Binary Vessel Dataset, Three-class Segmentation task on General Artery-Vein Dataset, and Nine-class Segmentation task on Fine-grained Artery-Vein Dataset.
Since the binary, general artery-vein and fine-grained vessel segmentation tasks are essentially semantic segmentation, we adopt the mean Intersection over Union (mIoU), mean Accuracy (mAcc), and mean F-score (mFscore) to evaluate the performance of models on our dataset.
Note that mFscore is of special interest since we have provided annotations for multi-class segmentation.
{(2) SVP recognition and temporal localization metrics:}
% Based on our temporal annotations, we can conduct SVP Recognition and SVP localization.
SVP recognition is to classify whether SVP exists in a video.
SVP localization task is to identify the time period where SVP appears in a video.
% Due to the clinical significance of ``SVP-absence'', it is crucial for us to detect ``SVP-absence'' cases as much as possible in SVP recognition task. 
% Therefore, we set the recall of ``SVP-absence'' as one of our evaluation metrics in SVP recognition.
% We employ the mean Average Recall (mAR) as a metric for our SVP localization task.
{We adopt the Accuracy (Acc), Area Under the Receiver Operating Characteristic Curve (AUROC), and Recall for SVP recognition.}
The frame-mAP (F-mAP), video-mAP (V-mAP)~\cite{vmap} under IoU threshold of 0.5 and mean Intersection over Union (mIOU) are adopted for the task of SVP localization.
%The videos are all 30 frames per second and range in duration from 1 second to 30 seconds. % To accurately determine the presence of SVP, clinicians need to observe at least one full clear cycle of the pulsation in a fundus video.
%As a result, video clips that contain continuous ODR with sufficient temporal duration (\ie, longer than a full cycle of SVP) are kept for further stabilization.

\newcommand\Tstrut{\rule{-2pt}{2.6ex}}         % = `top' strut
\newcommand\Bstrut{\rule[-0.9ex]{-2pt}{0pt}}   % = `bottom' strut
\newcommand{\hlcell}{{\cellcolor[rgb]{0.8,0.8,0.8}}}

% \vspace{-0.5em}
\section{Experiments}
% \vspace{-1em}
% To validate the challenge of RVD, we conduct extensive experiments on our dataset. We report the performance of multiple semantic segmentation methods and temporal segmentation methods based on variable annotations.
% Furthermore, to demonstrate the 
In this section, we employ state-of-the-art (SOTA) segmentation methods to examine the contributions and challenges of our newly curated RVD as well as establish a new benchmark for the dynamic vessel segmentation and localization tasks. As our data are collected from hand-held fundus imaging devices, we also investigate whether domain gaps 
between our dataset and existing ones which are captured by benchtop based devices.
% The domain shift between our dataset and existing datasets is further revealed in the experiments. %Overall, the results and analysis below underscore the unique contributions and potential advancements that our dataset could bring to the field.

%Then we report the performance of these benchmarks and provide a detailed analysis. Moreover, we explore the domain gaps between our dataset and previous ones. Such gaps reveal the distinctive attributes and challenges inherent to our dataset. Overall, the experimental results here help to reveal the uniqueness and potential advancements that our dataset can contribute to the field.

\begin{table}[t]
    % \footnotesize
    % \scriptsize
    \tiny
    \centering
    \caption{
        Segmentation results of different methods on our RVD dataset. ``DLV3'' denotes ``DeepLabV3'' and ``M2F'' denotes ``Mask2Former''.
        % We consider binary segmentation, general artery-vein segmentation, and fine-grained artery-vein segmentation.
    }
    \setlength{\tabcolsep}{0.35mm}{
        \begin{tabular}{lc|ccc|ccc|ccc}
            \toprule
            \multirow{2}{*}{Method}                                 & \multirow{2}{*}{Backbone}         & \multicolumn{3}{c|}{Binary}       & \multicolumn{3}{c|}{General Artery-Vein}  & \multicolumn{3}{c}{Fine-grained Artery-Vein} \\ 
            % \cline{3-11} 
                                         &                                                              & mIoU      & mAcc      & mFscore   & mIoU      & mAcc      & mFscore           & mIoU      & mAcc                               & mFscore            \\ 
            % \hline \hline
            \midrule \midrule
            % \multirow{3}{*}{FCN~\cite{shelhamer2017fully}}          & UNet~\cite{ronneberger2015u}      & 67.82     & 73.22     & 77.08     & 38.29     & 39.85     & 64.59             & 13.47     & 14.88                                                           & 68.95                          \\
            %                                                         & ResNet50~\cite{he2015deep}        & 62.12     & 66.05     & 70.76     & 49.22     & 53.93     & 58.99             & 18.38     & 21.41   & 26.62                               \\
            %                                                         & ResNet101                         & 62.79     & 66.77     & 71.54     & 48.24     & 51.98     & 57.90             & 18.53     & 21.44   &  24.07              \Bstrut         \\ 
            % \hline
            % \multirow{3}{*}{PSPNet~\cite{zhao2017pspnet}}           & UNet                              & 68.53     & 74.04     & 77.80     & 40.08     & 42.27     & 45.65             & 12.71     & 13.67                                                           & 64.34               \Tstrut         \\
            %                                                         & ResNet50                          & 61.82     & 65.25     & 70.37     & 49.08     & 53.94     & 58.71             & 18.92     & 22.10   & 24.45                               \\
            %                                                         & ResNet101                         & 63.06     & 67.11     & 71.87     & 47.76     & 51.34     & 57.12             & 19.37     & 22.39  & 25.05                \Bstrut         \\ 
            % \hline
            \multirow{3}{*}{DLV3~\cite{chen2017rethinking}}    & UNet~\cite{ronneberger2015u}                              & $66.59_{\pm 0.6}$     & $72.92_{\pm 1.0}$     & $76.67_{\pm 0.9}$     & $36.54_{\pm 1.5}$     & $37.80_{\pm 1.9}$     & $55.85_{\pm 1.2}$             & $12.81_{\pm 0.4}$     & $14.03_{\pm 0.7}$                                                           & $19.16_{\pm 0.5}$                \\
                                                                    & ResNet50~\cite{he2015deep}                          & $62.15_{\pm 0.6}$     & $65.84_{\pm 1.1}$     & $70.72_{\pm 0.7}$     & $47.92_{\pm 0.1}$     & $51.85_{\pm 0.2}$     & $57.21_{\pm 0.1}$             & $17.30_{\pm 0.3}$     & $19.59_{\pm 0.7}$   & $24.67_{\pm 0.5}$ \\
                                                                    & ResNet101                         & $62.89_{\pm 0.7}$     & $71.45_{\pm 1.0}$     & $78.45_{\pm 1.3}$     & $56.60_{\pm 0.7}$     & $51.99_{\pm 0.9}$     & $57.10_{\pm 1.0}$             & $18.13_{\pm 0.5}$     & $21.05_{\pm 0.6}$  & $24.72_{\pm 0.6}$                \Bstrut         \\ 
            % \hline
            % \multirow{4}{*}{Segmentor~\cite{strudel2021segmenter}}  & ViT-T~\cite{dosovitskiy2020image} & 49.39     & 51.25     & 51.51     & 33.83     & 35.15     & 36.04             & 11.98     & 12.57                                                           & 29.73               \Tstrut         \\ 
            %                                                         & ViT-S                             & 51.36     & 53.33     & 55.14     & 32.54     & 33.79     & 33.61             & 11.62     & 12.11   & 28.37                               \\
            %                                                         & ViT-B                             & 50.98     & 52.90     & 54.45     & 34.03     & 35.36     & 36.40             & 11.78     & 12.30   & 28.99                               \\
            %                                                         & ViT-L                             & 48.11     & 50.00     & 98.07     & 34.70     & 36.03     & 37.55             & 12.19     & 12.75   & 24.37               \Bstrut         \\
            \hline
            \multirow{7}{*}{M2F~\cite{cheng2021mask2former}}& ResNet50                          & $70.27_{\pm 0.3}$ & $77.65_{\pm 0.2}$ & $79.51_{\pm 0.3}$   & $57.60_{\pm 0.1}$    & $66.80_{\pm 0.2}$    & $69.06_{\pm 0.1}$     & $24.88_{\pm 0.8}$             & $32.58_{\pm 1.5}$            & $34.11_{\pm 1.1}$              \Tstrut \\
                                                                    & ResNet101                         & $70.74_{\pm 0.3}$ & $78.78_{\pm 0.3}$ & $79.99_{\pm 0.3}$   & $59.43_{\pm 1.7}$    & $68.58_{\pm 1.5}$    & $70.73_{\pm 1.7}$     & \hlcell $31.62_{\pm 4.0}$             & \hlcell $41.96_{\pm 5.1}$            & \hlcell $42.89_{\pm 6.4}$              \\
                                                                    & Swin-T~\cite{liu2021Swin}         & $70.94_{\pm 0.7}$ & \hlcell $78.87_{\pm 0.7}$ & $80.14_{\pm 0.7}$   & $58.58_{\pm 1.2}$    & $69.10_{\pm 2.6}$    & $71.39_{\pm 0.1}$     & $28.14_{\pm 3.2}$             & $36.93_{\pm 4.3}$            & $38.36_{\pm 4.2}$              \\ 
                                                                    & Swin-S                            & $70.27_{\pm 0.3}$ & $77.55_{\pm 0.1}$ & $80.14_{\pm 0.7}$   & $57.60_{\pm 0.1}$    & $66.14_{\pm 0.8}$    & $69.04_{\pm 0.1}$     & $23.41_{\pm 0.1}$             & $30.44_{\pm 0.5}$            & $32.60_{\pm 0.6}$              \\
                                                                    & Swin-B-1k                         & $71.20_{\pm 0.6}$ & $78.74_{\pm 0.9}$ & $80.38_{\pm 0.6}$   & $58.66_{\pm 0.3}$    & $68.43_{\pm 0.4}$    & $71.29_{\pm 0.8}$     & $25.30_{\pm 0.3}$             & $34.50_{\pm 1.0}$            & $34.81_{\pm 0.4}$              \\
                                                                    & Swin-B-22k                        & $70.99_{\pm 0.1}$ & $78.85_{\pm 0.6}$ & $80.19_{\pm 0.1}$   & $56.12_{\pm 1.3}$    & $68.31_{\pm 0.2}$    & $70.14_{\pm 0.3}$     & $25.26_{\pm 0.2}$             & $33.88_{\pm 0.3}$            & $34.88_{\pm 0.2}$              \\
                                                                    & Swin-L                            & \hlcell $74.09_{\pm 3.0}$ & $78.70_{\pm 0.9}$ & \hlcell $80.63_{\pm 0.4}$   & \hlcell $60.49_{\pm 1.7}$    & \hlcell $70.34_{\pm 1.9}$    & \hlcell $71.99_{\pm 1.6}$     & $24.91_{\pm 0.2}$             & $33.04_{\pm 0.7}$            & $34.46_{\pm 0.3}$              \Bstrut \\
            \bottomrule
        \end{tabular}
    }
    \label{tab: overall results of spatial segmentation}
    % \vspace{-1.5em}
\end{table}
%\vspace{-1.0em}
\subsection{Overall Results}
%\vspace{-0.75em}
\label{subsec: overall results}

%In examining the performance of existing SOTA models on our dataset, 

We first focus on spatial segmentation tasks on our dataset. We conduct experiments of binary vessel segmentation, general artery-vein segmentation, and fine-grained artery-vein segmentation, respectively. We employ several popular used segmentation methods, including FCN~\cite{shelhamer2017fully},
% PSPNet~\cite{zhao2017pspnet},
DeepLabV3~\cite{chen2017rethinking}, Segmentor~\cite{strudel2021segmenter}, and Mask2Former~\cite{cheng2021mask2former}. We also apply different backbones to these segmentation methods. The adopted backbones involves convolutional UNet~\cite{ronneberger2015u}, ResNet~\cite{he2015deep}, ViT~\cite{dosovitskiy2020image}, and Swin Transformer~\cite{liu2021Swin}. We use the pre-trained parameters as initialization and train the networks on our training set. 
Due to the space limit, we only present the results of DeepLabV3 and Mask2Former in Table~\ref{tab: overall results of spatial segmentation} and others are reported in the Appendix.
% \vspace{-0.55em}

\begin{figure}[t]
    \centering
    \includegraphics[width=0.9\textwidth]{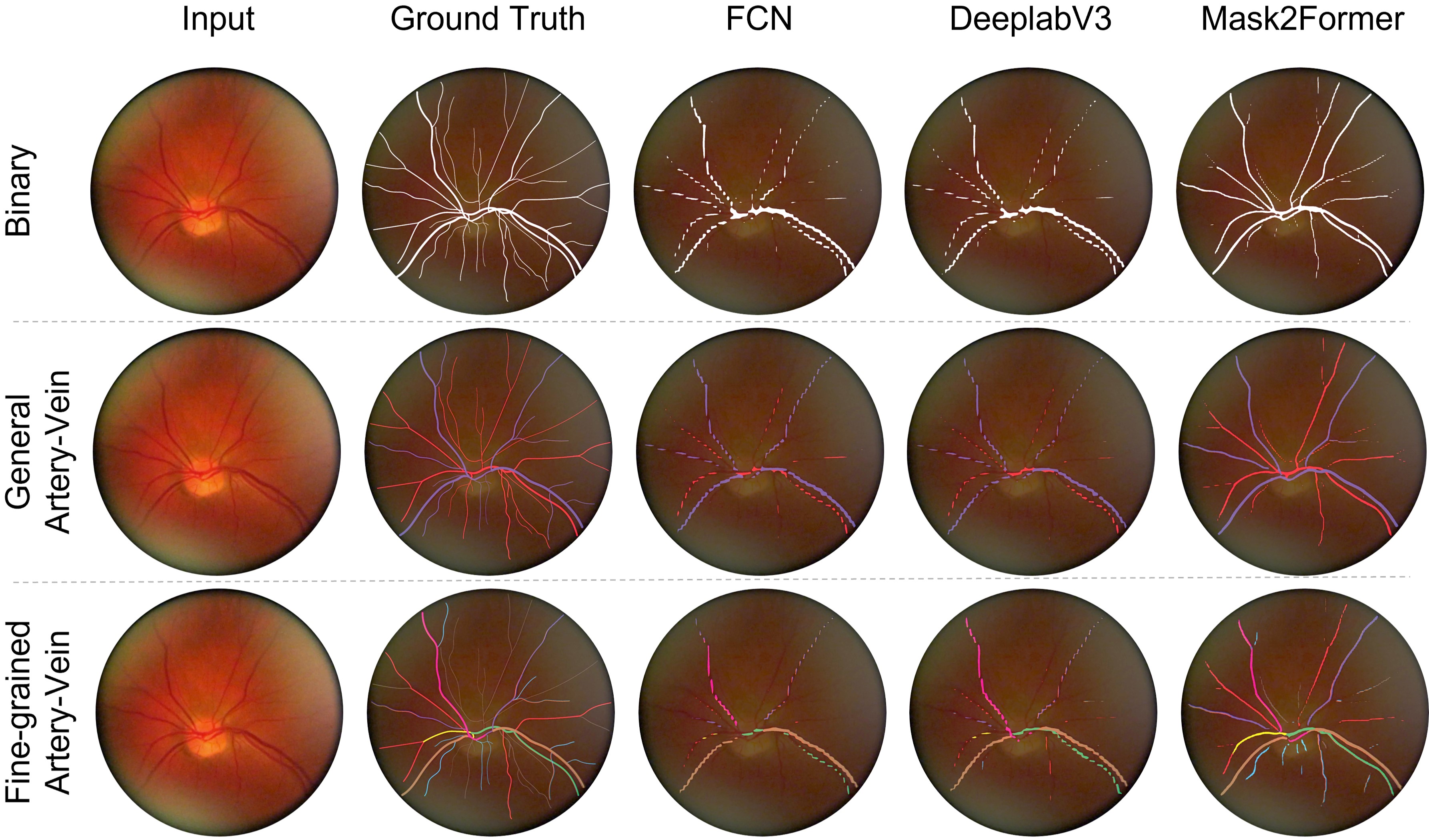}
    \caption{Visualization in the binary, general artery-vein, and fine-grained artery-vein segmentation.}
    % \vspace{-1.0em}
    \label{fig:case_study}
\end{figure}

%Generally, all the evaluated methods deliver the best performance in binary segmentation (with 1 class + background) and the worst results in fine-grained artery-vein segmentation (with 8 classes + background). A comparative evaluation across binary segmentation, general artery-vein segmentation, and fine-grained artery-vein segmentation reveals a significant decline in performance. Such results suggest that an increase in the number of classes intensifies the challenges of the segmentation tasks. 
Even for the binary segmentation task, the highest mIoU results barely reach around 70\%. For the more complex fine-grained artery-vein segmentation, the mIoU values further decline to approximately 25\%. Some visual results are shown in Fig.~\ref{fig:case_study}. It is observed that current methods in general struggle to localize thin vessels. However, these thin vessels always play an important role in reflecting some diseases, \eg, atherosclerosis~\cite{wong2006retinal}. %\XY{add one sentence saying small vessels are important for eye disease diagnosis in clinics. provide a referent as well}
% The performance of SOTA methods in our dataset is quite modest. 
The performance of the SOTA methods implies the challenges of our dataset, but this also emphasizes the potential of our dataset for future studies.
%More specifically, in comparison with DeepLabV3, Mask2Former consistently achieves better results across different backbones. Compared to the gain brought by different methods, the gain brought by different backbones used in each method is marginal.
%In the results of FCN, PSPNet, and DeepLabV3, backbone UNet outperforms ResNets and leads better results in binary segmentation (which involves 2 classes) and general artery-vein segmentation (which involves 3 classes). The reason could be that ResNets are more easily overfitted when the number of classes is limited. However, for fine-grained artery-vein segmentation with nine classes, ResNets tend to deliver better results. To some extent, this finding also explains why the UNet backbone is more commonly used than the ResNet backbone in retinal vessel segmentation research. %Moreover, it is noteworthy that Segmentor with a ViT-based backbone tends to yield comparatively lower results. We attribute this phenomenon to that ViTs may overlook the vessel details during the fine-tuning process.
%the lower sensitivity of ViTs to the details of vessels.

\begin{wraptable}{r}{9cm}
% \vspace{-1.25em}
    % \footnotesize
    % \scriptsize
    \scriptsize
    \centering
    \caption{
        Performance of SVP recognition and localization.
    }
    \setlength{\tabcolsep}{2.6mm}{
        \begin{tabular}{c|ccc|ccc}
            \toprule
            \multirow{2}{*}{Method} & \multicolumn{3}{c|}{Recognition} & \multicolumn{3}{c}{Localization} \\ %\cline{2-7} 
                                      & Acc & AUROC & Recall  & F-mAP & V-mAP & mIOU \\ 
            \midrule \midrule
            LRCN~\cite{lrcn}    & 52.68 & 56.79 & 45.00 & 64.62 & 59.06 & 50.62 \\
            I3D~\cite{i3d}      & 60.71 & 61.83 & 61.67 & 67.49 & 60.63 & 51.89 \\
            X3D~\cite{x3d}      & 52.68 & 51.60 & 75.00 & 61.12 & 52.60 & 50.53 \\
            TSN~\cite{tsn}      & 50.89 & 64.39 & 30.00 & 67.60 & 56.85 & 50.89 \\ 
            VTN~\cite{VTN}      & 58.93 & 65.58 & 86.67 & 68.08 & 57.64 & 51.25 \\
            \bottomrule
        \end{tabular}
        }
    \label{tab: temporal annotation for svp}
% \vspace{-1.8em}
\end{wraptable}
We also conduct experiments on temporal SVP recognition and localization with our provided annotations. 
% Specifically, we use the annotation to perform SVP recognition and SVP localization. 
In SVP recognition, we train the models to predict whether SVP exists in a video. 
In SVP localization, we train the models to identify the time period where SVP appears in a video. 
% In SVP localization, we train the models to identify the time period where SVP appears in a video clip.
We employ the models of LRCN~\cite{lrcn}, I3D~\cite{i3d}, X3D~\cite{x3d}, TSN~\cite{tsn}, and VTN~\cite{VTN}. We use the metrics in Section~\ref{subsec:protocol} and report the results in Table~\ref{tab: temporal annotation for svp}. 
To the best of our knowledge, we are the first to provide data and annotations for SVP recognition and localization. 
However, it is found that existing methods fail to recognize and localize SVP precisely on our real-clinic video data.  
% For example, in SVP localization task, despite the fact that SVP appears in most frames, VTN only achieves 51.25\% mIoU, which might not meet the needs of real-world applications. 
For example, in SVP localization task, VTN only achieves 51.25\% mIoU, which might not meet the needs of real-world applications. 
Such results indicate that more specific-designed methods are highly demanded.
%\vspace{-1em}
% designed with our proposed datasets in the future. 
% It is worthy to notice that the unsatisfying results of selected methods in both SVP recognition and SVP localization indicate a huge challenge in real-world fundus video applications.
%Overall, the results of selected methods  are not satisfactory, indicating a huge challenge in real-world fundus video applications. 
%We emphasize that we are the first to provide annotations for SVP recognition and localization. 
%Meanwhile, the precise detection of SVP is always required and is highly demanded by a variety of ocular diseases.
%Considering the absence of related datasets, we argue that our dataset contributes significantly in this area. We analyze three reasons to explain the low performance of adopted models. Normally, SVP only happens in only a small region in a frame. The existence of SVP could be inconsistent in video frames. Besides, the noise is imposed by hand-held devices.

% \begin{wrapfigure}{r}{0.35\textwidth}
%     \vspace{-5em}
%     \begin{center}
%         \includegraphics[width=0.34\textwidth]{Image/case_study.pdf}
%     \end{center}
%     \caption{
%         The segmentation masks predicted by Mask2Former. %(a) Masks generated in binary segmentation; (b) Masks generated in general artery-vein segmentation; (c) Masks generated in fine-grained artery-vein segmentation.
%     }
%     \label{fig:case_study}
%     \vspace{-3em}
% \end{wrapfigure}

% \vspace{-1em}
\subsection{Domain Gaps between RVD and Existing Datasets}
% \vspace{-0.75em}
We conduct a two-way evaluation process where models trained on our dataset are tested on previous datasets and models trained on existing datasets are also evaluated on our dataset. 
First, to evaluate binary vessel segmentation performance, we include the following datasets: CHASE DB1~(C-DB.), DRIVE~(DRI.), HRF, and STARE~(STA.). Then, we also conduct general artery-vein segmentation on the RITE dataset. The results are shown in Table~\ref{tab: domain gap with binary seg}. 
Note that existing datasets do not support fine-grained eight-class segmentation, and thus we did not test our data in this setting.
% \XY{Once we got the results, we can say something here. PUt figures as well!!!}
% The diminished performance of models when applied to a new dataset underscores the domain shift between our dataset and previous ones. 
Due to the domain gap, the models suffer performance drop.
The results also indicate that our dataset provides unique data samples. The visualization is illustrated in the Appendix. %Some predicted results are visualized in Fig.~\ref{fig: cross_dataset_case_study}.
From our experimental results above, we can tell that the retinal vessel segmentation is far from being solved. Our RVD dataset will serve as a valuable resource, motivating future explorations in retinal vessel segmentation.

% \begin{figure}
%     \centering
%     \includegraphics[width=\textwidth]{Image/cross_dataset_case_study.pdf}
%     \caption{Caption}
%     \label{fig: cross_dataset_case_study}
% \end{figure}

\begin{table}[t]
    % \footnotesize
    % \scriptsize
    \tiny
    \centering
    \caption{
         Evaluation of domain gaps between different datasets.
    }
    \setlength{\tabcolsep}{0.4mm}{
        \begin{tabular}{lc|cc|cc|cc|cc|cc}
            \toprule
            \multirow{2}{*}[-8pt]{Method}     & \multirow{2}{*}[-8pt]{Backbone} & \multicolumn{8}{c|}{\makecell[c]{Binary \\ Segmentation}}     & \multicolumn{2}{c}{\makecell[c]{General \\ Artery-Vein}} \Bstrut \\
                                      &         & \makecell[c]{RVD \\ $\downarrow$ \\ C-DB.} & \makecell[c]{C-DB. \\ $\downarrow$ \\ RVD} & \makecell[c]{RVD \\ $\downarrow$ \\ DRI.} & \makecell[c]{DRI. \\ $\downarrow$ \\ RVD} & \makecell[c]{RVD \\ $\downarrow$ \\ HRF} & \makecell[c]{HRF \\ $\downarrow$ \\ RVD} & \makecell[c]{RVD \\ $\downarrow$ \\ STA.} & \makecell[c]{STA. \\ $\downarrow$ \\ RVD}  & \makecell[c]{RVD \\ $\downarrow$ \\ RITE} & \makecell[c]{RITE \\ $\downarrow$ \\ RVD} \Tstrut \\ % & \makecell[c]{RVD \\ $\downarrow$ \\ WIDE} & \makecell[c]{WIDE \\ $\downarrow$ \\ RVD} & \makecell[c]{RVD \\ $\downarrow$ \\ DRI.} & \makecell[c]{DRI. \\ $\downarrow$ \\ RVD} & \makecell[c]{RVD \\ $\downarrow$ \\ HRF} & \makecell[c]{HRF \\ $\downarrow$ \\ RVD}  \Tstrut \\ 
            % \cline{3-11} 
            % \hline \hline
            \midrule \midrule
            % \multirow{3}{*}{FCN}         & UNet      &          &       &       &       &       &       &       &       &       &       \\
            %                              & ResNet50  &          &       &       &       &       &       &       &       &       &       \\
            %                              & ResNet101 &          &       &       &       &       &       &       &       &       &       \Bstrut     \\
            % \hline
            % \multirow{3}{*}{PSPNet}      & UNet      &          &       &       &       &       &       &       &       &       &       \Tstrut     \\
            %                              & ResNet50  &          &       &       &       &       &       &       &       &       &       \\
            %                              & ResNet101 &          &       &       &       &       &       &       &       &       &       \Bstrut     \\
            % \hline
            \multirow{3}{*}{DLV3}   & UNet         & $70.23_{\pm 0.2}$ & $62.17_{\pm 0.4}$ & $65.21_{\pm 0.2}$ & $62.21_{\pm 0.3}$ & $70.89_{\pm 0.3}$ & $56.81_{\pm 0.5}$ & $70.58_{\pm 0.3}$ & $62.47_{\pm 0.2}$  & $51.05_{\pm 0.2}$ & $28.31_{\pm 0.4}$  \\
                                         & ResNet50     & $71.56_{\pm 0.3}$ & $63.41_{\pm 0.2}$ & $66.34_{\pm 0.1}$ & $63.04_{\pm 0.2}$ & $71.27_{\pm 0.2}$ & $57.76_{\pm 0.4}$ & $71.23_{\pm 0.2}$ & $63.13_{\pm 0.1}$    & $51.84_{\pm 0.3}$ & $29.87_{\pm 0.3}$ \\
                                         & ResNet101    & $71.89_{\pm 0.4}$ & $63.59_{\pm 0.2}$ & $66.72_{\pm 0.3}$ & $63.98_{\pm 0.1}$ & $71.98_{\pm 0.3}$ & $58.17_{\pm 0.9}$ & $71.98_{\pm 0.4}$ & $63.89_{\pm 0.3}$ & $52.56_{\pm 0.2}$ & $30.30_{\pm 0.4}$ \Bstrut     \\
            \hline
            % \multirow{4}{*}{Segmentor}   & ViT-T     &          &       &       &       &       &       &       &       &       &       \Tstrut     \\
            %                              & ViT-S     &          &       &       &       &       &       &       &       &       &       \\
            %                              & ViT-B     &          &       &       &       &       &       &       &       &       &       \\
            %                              & ViT-L     &          &       &       &       &       &       &       &       &       &       \Bstrut     \\
            % \hline
            \multirow{7}{*}{M2F} & ResNet50     & $72.23_{\pm 0.2}$ & $65.71_{\pm 0.3}$ & $67.03_{\pm 0.2}$ & $66.74_{\pm 0.1}$ & $73.05_{\pm 0.2}$ & $65.65_{\pm 0.2}$ & $73.79_{\pm 0.3}$ & $64.00_{\pm 0.2}$ & $54.65_{\pm 0.2}$ & $49.84_{\pm 0.3}$ \Tstrut     \\
                                         & ResNet101    & $73.57_{\pm 0.3}$ & $67.20_{\pm 0.1}$ & $67.57_{\pm 0.2}$ & $67.25_{\pm 0.3}$ & $73.76_{\pm 0.3}$ & $65.66_{\pm 0.1}$ & $74.31_{\pm 0.2}$ & $64.80_{\pm 0.4}$ & $55.23_{\pm 0.3}$ & $50.12_{\pm 0.4}$ \\
                                         & Swin-T       & $74.97_{\pm 0.2}$ & $67.59_{\pm 0.3}$ & $69.99_{\pm 0.3}$ & $67.74_{\pm 0.3}$ & $74.71_{\pm 0.1}$ & $63.89_{\pm 0.3}$ & $74.73_{\pm 0.2}$ & $65.10_{\pm 0.2}$ & $55.65_{\pm 0.4}$ & $50.88_{\pm 0.2}$ \\
                                         & Swin-S       & $75.21_{\pm 0.9}$ & $67.98_{\pm 0.2}$ & $69.93_{\pm 0.1}$ & $67.51_{\pm 0.3}$ & $73.97_{\pm 0.2}$ & $67.05_{\pm 0.3}$ & $75.12_{\pm 0.3}$ & \hlcell $67.50_{\pm 0.3}$ & $56.47_{\pm 0.2}$ & $51.42_{\pm 0.3}$ \\
                                         & Swin-B-1k    & $74.93_{\pm 0.2}$ & $68.13_{\pm 0.1}$ & $71.21_{\pm 0.2}$ & $67.64_{\pm 0.3}$ & $76.03_{\pm 0.3}$ & $66.88_{\pm 0.4}$ & $74.70_{\pm 0.1}$ & $67.21_{\pm 0.3}$ & $57.04_{\pm 0.4}$ & $52.99_{\pm 0.4}$ \\
                                         & Swin-B-22k   & \hlcell $76.95_{\pm 0.3}$ & $70.04_{\pm 0.2}$ & \hlcell $73.79_{\pm 0.3}$ & \hlcell $68.36_{\pm 0.3}$ & \hlcell $78.67_{\pm 0.2}$ & $64.06_{\pm 0.4}$ & \hlcell $78.28_{\pm 0.3}$ & $65.95_{\pm 0.3}$ & $57.05_{\pm 0.2}$          & $52.84_{\pm 0.5}$ \\
                                         & Swin-L       & $76.86_{\pm 1.3}$ & \hlcell $70.47_{\pm 0.2}$ & $73.02_{\pm 0.1}$ & $67.72_{\pm 0.4}$ & $76.15_{\pm 0.3}$ & \hlcell $67.11_{\pm 0.3}$ & $72.77_{\pm 0.4}$ & $65.88_{\pm 0.3}$ & \hlcell $57.28_{\pm 0.4}$ & \hlcell $52.98_{\pm 0.3}$ \Bstrut     \\
            \bottomrule
        \end{tabular}
    }
    \label{tab: domain gap with binary seg}
    %\vspace{-3.0em}
\end{table}

\vspace{-1.0em}
\section{Conclusion}
\vspace{-1.0em}
In this work, we propose the first video-based retinal vessel segmentation dataset by employing handheld devices for data acquisition.
% These annotations correlate significantly with a range of clinical objectives. 
Our dataset significantly complements the current benchtop-based datasets for retinal vessel segmentation and enables SVP detection and localization.
More importantly, it offers rich annotations for both spatial vessel segmentation and temporal SVP localization. 
% the distribution of videos in our dataset is closer to that in real-world scenarios. 
In comparison to existing datasets, our dataset is not only the largest scale one with the most diverse annotations but also more challenging. The domain gaps between our dataset and existing ones allow researchers to investigate how to minimize the domain gaps in vessel segmentation. 
Therefore, our curated dataset RVD is valuable for retinal vessel segmentation and would facilitate the clinical diagnosis of eye-related diseases.
\clearpage
{\small
\bibliographystyle{plainnat}
\bibliography{neurips_data_2023}
}
\end{document}